# The Transformer Paradigm: Dynamic Processing and a New Principle of Inference[1]

**Marco Giunti**
ALOPHIS – Applied Logic Philosophy and History of Science, University of Cagliari
ORCID: 0000-0003-4182-4850
Email: giunti@unica.it

**Fabrizia Giulia Garavaglia**
Dipartimento di Pedagogia, Psicologia, Filosofia, University of Cagliari
ORCID: 0009-0003-4699-4709
Email: fabriziag.garavaglia@unica.it

**Abstract.** We argue that the Transformer revolution is not adequately understood as a merely quantitative expansion of deep learning, but as the emergence of a new paradigm based on a different principle of inference. We interpret the Transformer as a system in which token representations are transformed by parameterized transformations, or transforming concepts. During inference, some of these transformations are dynamically constructed from the current input: previously acquired transformations are first adapted and then applied. At the mechanical-structural level, this dynamic processing is realized through output-weight interconnections, whereby outputs of some networks determine the weights of others. The resulting mode of processing, which we call Sequence-level Interactive Dynamic Parallel Processing (SIDPP), provides a framework for explaining strong prompt sensitivity and opens research directions concerning interpretability, control, alternative AI architectures, neuromorphic systems, and possible biological realizations of dynamic processing.

## 1 Introduction — The Transformer as a New Paradigm in AI

One long-standing challenge of artificial intelligence has been to build systems capable of open-ended natural-language conversation: not merely producing grammatical sentences, but responding appropriately across changing contexts and topics while drawing on shared background knowledge. This capacity was already central to Turing's imitation game, in which a machine had to sustain unrestricted linguistic interaction well enough to be mistaken for a human interlocutor (Turing 1950). For decades, however, it remained largely beyond the reach of the major paradigms of artificial intelligence.

[1] This article develops ideas and an interpretation of the Transformer presented by Marco Giunti at the conference *Exploring the Frontiers of Science with AI and the Role of Humanities*, promoted by RESILIENCE – The Religious Studies Research Infrastructure, Reggio Emilia, 9–11 March 2026.

The situation changed with the Transformer architecture introduced by Vaswani et al. (2017) and with the large language models subsequently built from it. The original Transformer did not yet display the capabilities of present systems, but it introduced an architectural template whose successive development, scaling, and adaptation made this form of linguistic interaction increasingly effective. Transformer-based systems have since become a central platform for a rapidly expanding family of models, applications, and research programs.

This development can be described, in a limited but substantive Kuhnian sense, as the emergence of a new paradigm in AI. The Transformer provided an architectural exemplar: a sufficiently novel and successful solution to attract a rapidly growing research community, yet sufficiently open-ended to generate further developments, applications, and problems (Kuhn 1970). This does not imply a complete rupture with deep learning, but the emergence within it of a new organizing model for research and technological development.

What conceptual and structural discontinuity, then, underlies this shift? We argue that it lies in a new mode of inference. The Transformer does not merely apply transformations fixed through training: during inference, previously acquired knowledge is adapted to the current input through dynamically constructed transformations and only then applied to token representations. At the mechanical-structural level, this dynamic processing is made possible by output-weight interconnections, through which the outputs of some networks determine the weights—and thus the transformations—of others.

The philosophical significance of this interpretation is that it changes the explanatory picture of Transformer inference. What the system does cannot be understood simply as the application of transformations acquired during training. Training also establishes an organization through which new, input-dependent transformations are constructed during inference. We therefore argue that the Transformer shift is not merely a quantitative expansion of the deep-learning paradigm, but the introduction of a different principle of processing.

The argument proceeds in two stages. Sections 2–4 develop the technical-conceptual thesis. Section 2 distinguishes transformee from transforming concepts and, within the latter, static from dynamic transformations. Section 3 provides a mechanical-structural interpretation of these transformations and identifies output-weight interconnections as the mechanism that makes dynamic processing possible. Section 4 characterizes the resulting organization as Sequence-level Interactive Dynamic Parallel Processing (SIDPP). Section 5 then uses GPT-4.5's recent Turing-test results as a behavioral illustration of strong prompt sensitivity (Jones and Bergen 2026a). On this basis, Section 6 explores some research perspectives opened by this interpretation, while Section 7 returns to the philosophical thesis: the Transformer revolution is not adequately understood as a merely quantitative expansion of deep learning, but as the emergence of a different principle of inference.

## 2 The Transformer as a System That Transforms Concepts by Means of Concepts

To identify the new principle of inference outlined above, we first need to distinguish what is transformed in the Transformer, what performs those transformations, and how the relevant transformations are determined.

### 2.1 Transformee Concepts

In the Transformer, what is transformed are vector representations of tokens. These include both the initial embedding vectors and the representations generated at successive stages of processing. We call them transformee concepts.

A token vector is a distributed representation: each of its positions corresponds to a representational feature, or property, while the numerical value occupying that position gives the value of that feature for the represented token (Hinton 1986; Bengio, Courville, and Vincent 2013). The representation of a token is therefore not localized in a single unit, but distributed across an ordered set of feature values.

Calling these representations concepts does not imply that their constituent features must correspond to concepts or categories recognizable by human interpreters. Individual dimensions may have no direct interpretation in ordinary semantic terms, and the same vector position need not correspond to the same feature or property at successive stages of processing (cf. Jawahar, Sagot, and Seddah 2019). What matters here is that, at each stage, the vector represents the token through an ordered set of feature values that can in turn be transformed by subsequent operations.

### 2.2 Transforming Concepts

Token representations are transformed by parameterized transformations. A matrix or vector of model parameters determines how an input representation is mapped onto a new representation. In previous work, we called such transformations transforming concepts (Giunti and Garavaglia 2026; hereafter TR1). The transforming concept is the transformation itself, not simply the matrix or vector that parameterizes it.

A central case is a linear transformation $\mathcal{L}[\boldsymbol{W}]$ defined by a parameter matrix $\boldsymbol{W}$. Applied to a token representation $\boldsymbol{v}$, it produces a new representation $\boldsymbol{v}' = \mathcal{L}[\boldsymbol{W}](\boldsymbol{v})$, whose values result from weighted combinations of the feature values in $\boldsymbol{v}$. Parameter vectors can analogously define componentwise transformations, such as addition or multiplication. In each case, the parameter structure determines how one structured representation is transformed into another.

The term *concept* does not imply that the transformation, or the features it operates on, must be interpretable in human conceptual categories. It indicates instead its functional role: a transforming concept specifies a way of transforming a transformee concept. From this perspective, the Transformer can be characterized as a system in which concepts are transformed by means of other concepts.

### 2.3 Static and Dynamic Transforming Concepts

Transforming concepts differ according to how their parameters are determined. Static transforming concepts are defined by matrices or vectors whose values are fixed through training. They therefore embody transformations acquired before inference and, in this limited sense, encode previously acquired knowledge. Once training is complete, these transformations remain fixed while the system processes a particular input.

During inference, however, the Transformer also generates matrices and vectors whose values depend on the current input. These structures in turn parameterize further transformations of token representations. We call the resulting transformations dynamic transforming concepts (TR1). Unlike static transforming concepts, they are therefore not fixed in advance of the current processing episode, but are defined by parameter structures constructed by applying appropriate static transforming concepts to the representations of the current input.

This yields a two-stage organization of inference. Static transforming concepts first contribute to the construction of input-dependent parameter structures; the dynamic transforming concepts thereby defined are then applied to token representations. In this sense, dynamic transforming concepts are input-dependent adaptations of static transforming concepts. Transformer inference thus involves not only the application of previously acquired transformations, but first their adaptation to the current input and then the application of the resulting transformations. Figure 1 summarizes the distinction between transformee concepts and static and dynamic transforming concepts.

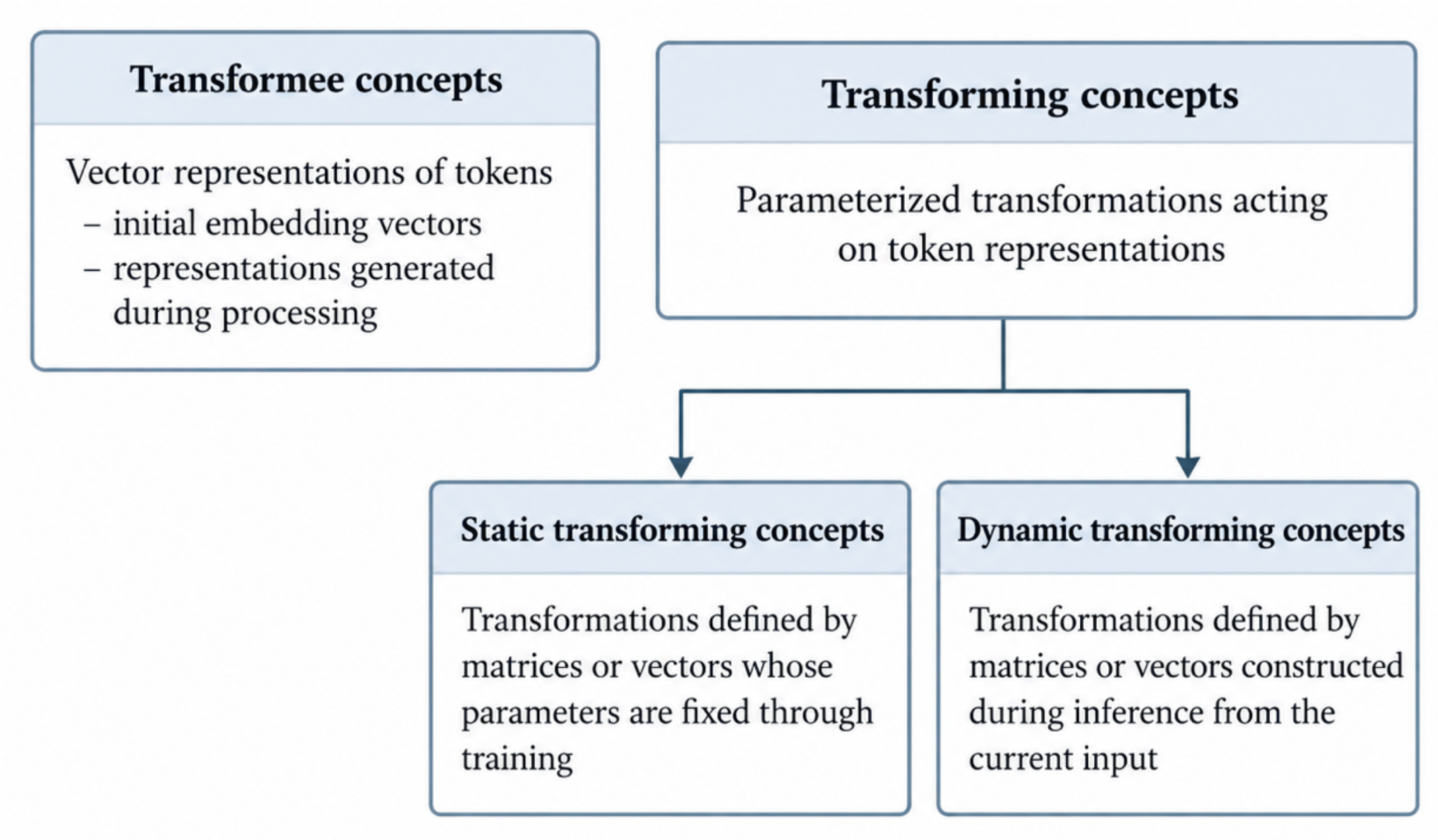

**Figure 1.** Transformee and transforming concepts in the Transformer. Transforming concepts are static when their parameters are fixed through training and dynamic when their defining parameter structures are constructed during inference from the current input.

This interpretation is consistent with work showing that large language models can exhibit learning-like adaptation during inference and in-context learning (Brown et al. 2020; Akyürek et al. 2023). Our claim, however, is more specific: it concerns the architectural mechanism by which input-dependent transformations are constructed and used during inference, rather than in-context learning as such.

## 3 From Transforming Concepts to a Mechanical-Structural Interpretation

The distinction between static and dynamic transforming concepts characterizes Transformer inference at the functional level. To understand how dynamic transformations are actually realized, we must now redescribe the same processing in terms of the components that perform these transformations and of the relations that organize them.

### 3.1 Transforming Concepts as Simple Neural Networks

Each transforming concept—that is, each parameterized transformation introduced in §2—can be identified, at the mechanical-structural level, with a simple neural network of the corresponding type. Consider the central case introduced in §2.2. A linear transformation $\mathcal{L}[\boldsymbol{W}]$, defined by a parameter matrix $\boldsymbol{W}$, can be identified with a dense linear network whose connection weights are the elements of $\boldsymbol{W}$. The network thereby realizes the transformation, mapping an input vector $\boldsymbol{v}$ onto the output $\boldsymbol{v}' = \mathcal{L}[\boldsymbol{W}](\boldsymbol{v})$. Figure 2 illustrates the corresponding mechanical-structural representation for the central case of a dense linear network with static weights.

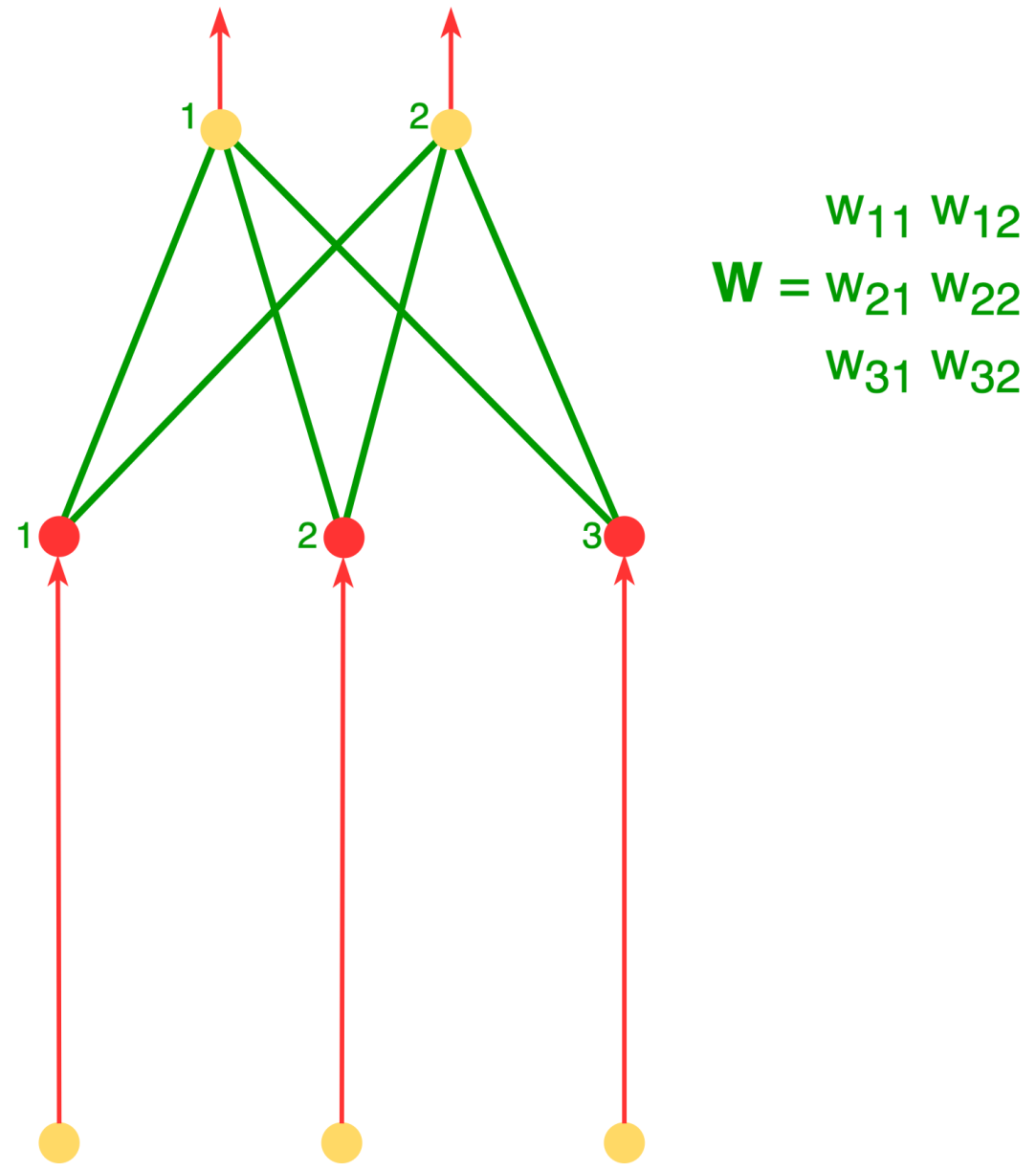

**Figure 2.** Traditional representation of a dense linear network with static weights. Output units compute the weighted sums of the signals received from input units, which merely route the incoming signals. Adapted from TR1, supplementary Fig. S17.

The same relation holds for transforming concepts defined by parameter vectors. Componentwise addition and multiplication can be identified with one-to-one additive and multiplicative networks, respectively, in which each input unit is connected only to the corresponding output unit. The two network types have the same basic one-to-one morphology and differ in the operation performed at their output units. In each case, the network realizes the parameterized transformation with which it is identified.

This identification does not replace the standard mathematical description of the Transformer. It redescribes the same transformations at a level at which the components realizing them and their organization can be made explicit. In mechanistic terms, this is the first step from a functional to a mechanical-structural description: transforming concepts can now be treated as identifiable components whose interconnections can be analyzed (Bechtel and Abrahamsen 2005; Kaplan and Craver 2011).

### 3.2 From Individual Networks to Groups of Networks

Transformer processing does not ordinarily apply a transforming concept to a single token representation in isolation. Rather, vector transformations are applied in parallel to the representations of the tokens in the current input sequence. In the standard mathematical description, this parallel organization is implemented through operations on matrices whose rows are those token representations, so that the resulting matrix contains the corresponding transformed representations as its rows.

The identification established in §3.1 can consequently be extended from individual transforming concepts to these matrix operations. Since each parameterized vector transformation can be identified with a simple neural network, a matrix operation that applies such transformations in parallel can be identified, at the mechanical-structural level, with a group of simple neural networks operating in parallel. Each active network processes one token representation and produces the corresponding representation at the next stage.

This shift from individual networks to groups of networks introduces a higher level of organization. What matters is no longer only which transformation each network realizes, but also how the groups are interconnected and how processing in one group can affect processing in another. The crucial case for present purposes is the one in which outputs of some networks become the parameters of others. We turn to these output-weight interconnections in §3.3.

### 3.3 Output-Weight Interconnections and Dynamic Transformations

The higher-level organization introduced in §3.2 becomes especially important when the output of one group of networks is not used as the input of another group, but instead provides its parameters. In the mechanical-structural description developed in TR1, this relation is represented by an output-weight interconnection: an output produced by one network becomes the weight of a connection in another network. The receiving network therefore does not operate with weights fixed in advance, but with weights whose values depend, during inference, on the current input. Figure 3 illustrates an output-weight interconnection in which outputs of other networks determine the dynamic weights of a dense linear network.

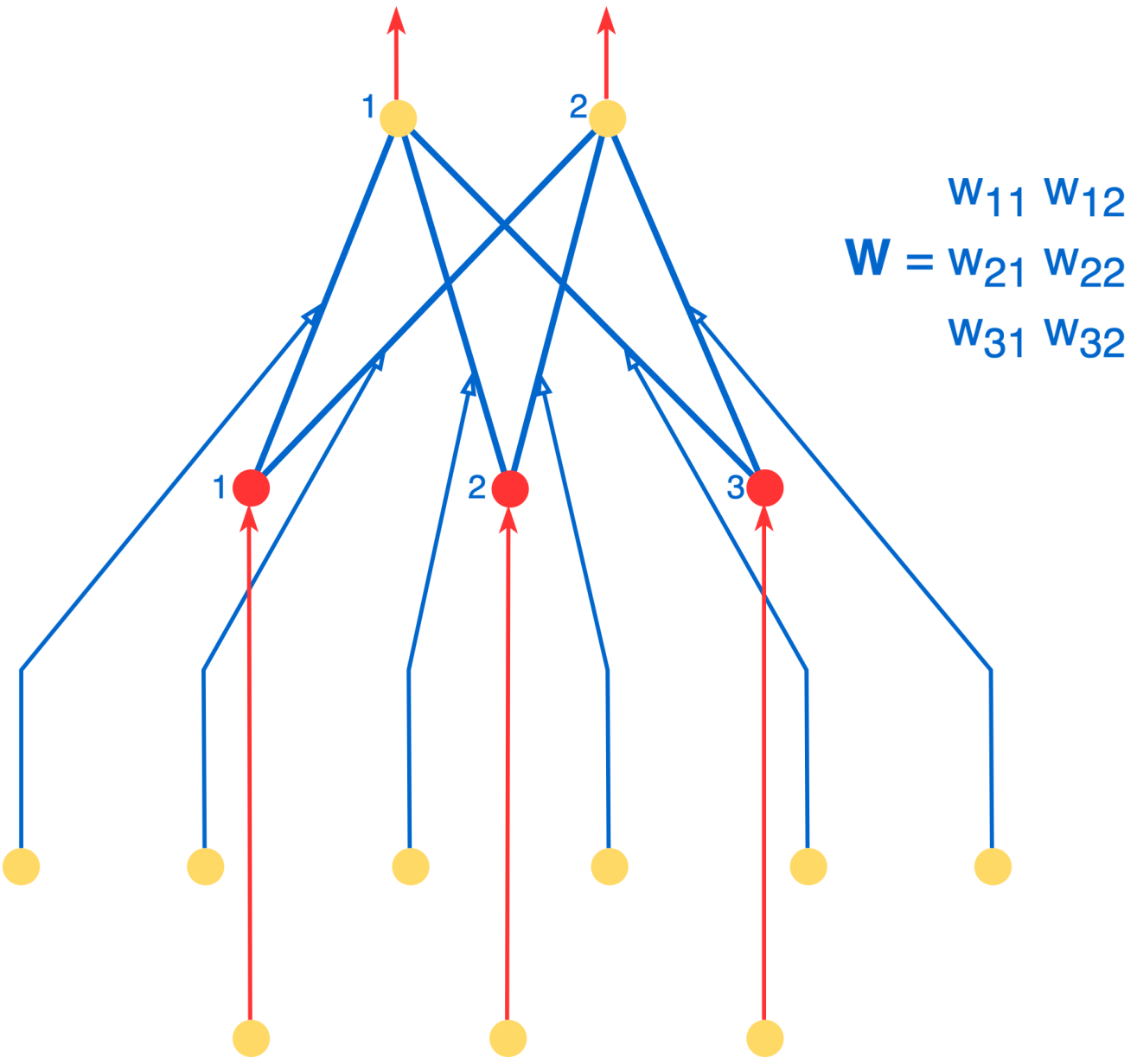

**Figure 3.** A dense linear network with dynamic weights determined by output-weight interconnections originating from output units of other networks. Adapted from TR1, Fig. 2.

Output-weight interconnections provide the mechanical-structural realization of the dynamic transforming concepts introduced in §2.3. Appropriate static transforming concepts first generate input-dependent parameter structures; through output-weight interconnections, the values in those structures become weights of other networks and thereby define the transformations those networks realize. The conceptual sequence identified earlier can thus be redescribed mechanically as follows: static transformations generate input-dependent weights, and the resulting transformations are then applied to token representations.

Generated or rapidly changing weights are not unique to the Transformer: related ideas appear in hypernetworks and fast-weight architectures (Ha, Dai, and Le 2017; Schlag, Irie, and Schmidhuber 2021). Our claim is therefore not that the Transformer introduced dynamic weights as such. Rather, the point is that the standard Transformer can be systematically redescribed as an architecture in which output-weight interconnections are constitutive of its ordinary inference processing. It is this organization that makes the dynamic processing identified in §2.3 mechanically explicit.

## 4 From Parallel Distributed Processing to Sequence-Level Interactive Dynamic Parallel Processing

The mechanical-structural description developed in §3 reveals a level of parallel organization that differs from the one traditionally emphasized in connectionist models. In classical Parallel Distributed Processing, parallelism primarily concerns the simultaneous activity of many units within a neural network, whose joint activation transforms an input vector representation into an output vector representation (Rumelhart, McClelland, and the PDP Research Group 1986). In the Transformer, by contrast, the relevant parallelism considered here occurs at a higher level: groups of simple neural networks operate in parallel on the vector representations of the tokens in an input sequence.

This parallel processing is sequence-level because the elements processed in parallel are the representations of the tokens composing the sequence. As described in §3.2, a matrix operation represents this mathematically by applying the relevant vector transformations in parallel to the rows of a matrix. In the corresponding mechanical-structural description, each active network in a group processes one token representation, while the group as a whole processes the sequence. Sequence-level parallelism is therefore not simply the parallel activity internal to an individual network, but a property of the organization of multiple networks processing corresponding elements of a sequence.

Sequence-level parallel processing in the Transformer is also interactive. The representation produced for one token need not depend only on the previous representation of that same token: through attention, it can depend on representations of other tokens in the current sequence. The parallel streams are therefore not independent. Their interaction is structurally organized so that information associated with different tokens can contribute to the transformations applied to a given token representation.

Finally, the processing is dynamic in the sense introduced in §§2.3 and 3.3. Some transformations applied during inference are not fixed by training alone. Their parameters depend on the current input and are supplied through output-weight interconnections. Thus the sequence being processed contributes not only information to be transformed, but also to the construction of some of the transformations through which that processing proceeds. Dynamic neural networks in a broader sense have precedents and constitute an established research area (Han et al. 2022); the present claim concerns the way these three properties are jointly realized in the standard Transformer architecture. Figure 4 shows how these three properties are jointly realized in a bidirectional attention head.

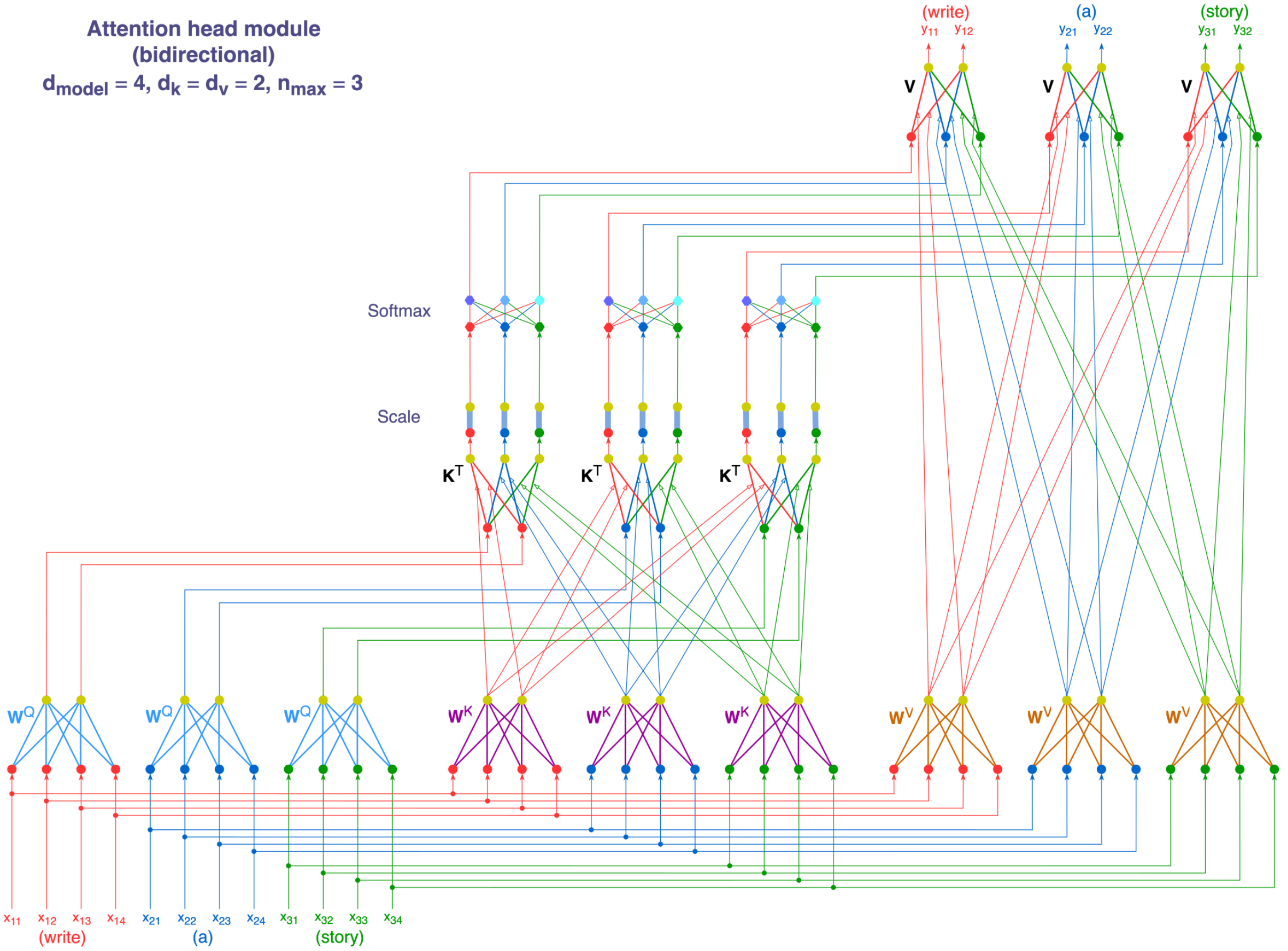

**Figure 4.** Bidirectional attention head with a 3-token input. The figure illustrates SIDPP: token representations are processed in parallel at the sequence level. Processing is dynamic because some of the transformations applied during inference are defined by parameter structures constructed from the current input through output-weight interconnections. These output-weight interconnections are also interwoven across token-processing streams, so that the representation of each token can depend on those of the others; processing is thus interactive. Adapted from TR1, supplementary Fig. S11c.

We call this mode of processing Sequence-level Interactive Dynamic Parallel Processing (SIDPP). The term identifies a processing principle rather than a particular functional organization: the Transformer is one architecture that realizes it, but other organizations could in principle do so as well. Its three components pick out mutually connected aspects of the same mode of processing: token representations are processed in parallel at the sequence level; processing streams interact across tokens; and some of the transformations applied to those representations are dynamically constructed from the current input.

SIDPP therefore marks a level of organization not captured by describing the Transformer simply as a larger or deeper instance of Parallel Distributed Processing. The continuity with the connectionist tradition is substantial: processing remains distributed, parallel, and implemented through networks whose parameters are largely acquired through training. The discontinuity concerns the organization of inference. Parallel processing is coordinated across a sequence of representations, interaction among those representations is built into the architecture, and input-dependent transformations are constructed and applied during processing. In this sense, SIDPP provides the processing-level characterization of the discontinuity anticipated in §1.

## 5 Dynamic Processing in Action: GPT-4.5 and the Turing Test

The dynamic character of SIDPP makes Transformer processing highly sensitive to the current input. A particularly striking behavioral illustration is provided by the recent performance of GPT-4.5 in a standard three-party Turing test (Jones and Bergen 2026a). In this test, a human interrogator converses with two respondents, one human and one artificial, and must identify which is which on the basis of their responses.

Jones and Bergen tested GPT-4.5 under two system-prompt conditions in both a Prolific sample and a UCSD student sample. In the PERSONA condition, a 187-word prompt specified a personality and a detailed linguistic style; in the NO-PERSONA condition, a 21-word prompt merely instructed the model to convince the interrogator that it was human (Jones and Bergen 2026b). Table 1 reports the results aggregated across the two populations.

**Table 1. GPT-4.5 Turing-test results with PERSONA and NO-PERSONA prompts**

| Prompt condition | Prompt words | Judged human | Not judged human | Total | Success rate | Exact one-sided $p$ vs. .50 |
|---|---|---|---|---|---|---|
| PERSONA | 187 | 185 | 69 | 254 | 72.8% | < .001 |
| NO-PERSONA | 21 | 50 | 91 | 141 | 35.5% | < .001 |

*Note.* Data aggregated across the Prolific and UCSD student samples reported by Jones and Bergen (2026a); prompt lengths from Jones and Bergen (2026b). Exact one-sided binomial $p$-values against .50 were calculated from the aggregated counts: right-tailed for PERSONA and left-tailed for NO-PERSONA.

The contrast is substantial. With PERSONA, GPT-4.5 was judged human in 72.8% of the tests across the two populations, significantly above the .50 chance level; with NO-PERSONA, the corresponding rate was 35.5%, significantly below it. The PERSONA prompt was only 166 words longer, yet the aggregate success rate differed by 37.4 percentage points. The two prompt conditions therefore produced sharply different behavioral outcomes from the same trained model. The corresponding results are also substantial when the two populations are considered separately.

The framework developed in §§2–4 provides a plausible explanation of how such a striking reversal can result from a relatively small change in the information supplied by the prompt. The prompt does not merely add information to a processor whose transformations remain fixed. In each attention head (see Figure 4), specific static transforming concepts are applied to the representations of the current input to construct the parameter structures that define dynamic transforming concepts: in particular, the transformations defined by $\boldsymbol{W}^K$ and $\boldsymbol{W}^V$ generate $\boldsymbol{K}$ and $\boldsymbol{V}$, from which the dynamic transformations defined by $\boldsymbol{K}^\top$ and $\boldsymbol{V}$ are obtained.[2] Changing the prompt therefore changes the transformations subsequently applied during inference. When prompt length changes, as in the present case, some dynamically generated parameter matrices (namely, $\boldsymbol{K}^\top$ and $\boldsymbol{V}$) also change their dimensions, so that the corresponding transformations involve different numbers of dynamically generated parameters and, in the mechanical-structural description, different numbers of effective weights and connections. The system thus adapts its processing to the current input by constructing different input-dependent transformations, and these differences can propagate through subsequent processing to produce markedly different behavior.

[2] The two weight matrices $\boldsymbol{W}^K$ and $\boldsymbol{W}^V$ are parameter matrices of each attention head whose values are fixed through training. Applying the corresponding static transforming concepts to the current input produces the key matrix $\boldsymbol{K}$ and the value matrix $\boldsymbol{V}$; the matrix $\boldsymbol{K}^\top$ is the transpose of the key matrix.

In previous work, we used the term strong prompt sensitivity for this pronounced dependence of Transformer processing on the information supplied by the prompt (TR1). The GPT-4.5 results provide a particularly clear behavioral illustration of this phenomenon: a relatively small change in the prompt is associated with a large change in behavior, consistently with the view that prompt-dependent transformations are a constitutive part of Transformer inference. The example is therefore not offered as a test of the framework, but as an indication of the kind of behavioral effect that dynamic processing can plausibly help explain.

## 6 New Research Perspectives

The interpretation developed in the preceding sections suggests several research directions that go beyond the explanatory aims of the present paper. We consider three of them: the analysis and control of Transformer processing, the design of alternative architectures that realize dynamic processing more directly, and the possibility of analogous mechanisms in biological neural systems.

### 6.1 Interpretability, Predictability, and Control

The mechanical-structural interpretation developed in §3 makes explicit a class of dependencies that is less salient in the standard matrix description of the Transformer. During inference, representations of the current input contribute to the construction of dynamic parameters; these parameters define dynamic transformations; and the resulting transformations contribute to subsequent representations and, ultimately, to the system's output. The relevant explanatory chain can therefore be followed from input representations to dynamically constructed transformations and from these to later stages of processing.

This perspective may contribute to *interpretability* by shifting part of the analysis from the trained model considered as a fixed structure to the input-dependent transformations actually constructed during a particular processing episode. It does not make the Transformer transparent in every respect, nor does it provide a complete causal explanation of individual outputs. It does, however, identify additional internal dependencies that can be investigated alongside learned weights, features, and circuits (Creel 2020; Elhage et al. 2021).

The same dependencies are relevant to *control* and *predictability*. If changes in the prompt induce different dynamic transformations, one research strategy is to study which interventions on the input construct, stabilize, or suppress particular classes of transformations and how these changes propagate through subsequent processing. Predictability may likewise require combining global knowledge of the trained model with analysis of the dynamic transformations constructed for a particular input. In this sense, the framework suggests treating understanding not only as a description of fixed internal structure, but also as the identification of dependencies within the processing actually produced in a given context (cf. Dellsén 2020).

## 6.2 New AI Architectures: Neuromorphic Systems, Energy Efficiency, and Smaller Models

The mechanical-structural description also raises a design question. In present Transformer systems, the organization described in §3 is implemented through large-scale matrix operations executed on conventional digital hardware. But the functional principle itself does not require this particular realization. Output-weight interconnections could in principle be realized more directly by mechanisms in which the output of one component locally modifies the transformation performed by another. This suggests a research program aimed at architectures that implement dynamic transformations without reproducing the same organization through massive matrix computation.

Neuromorphic computing provides one possible direction. Its general aim is to realize information processing through hardware whose organization more directly reflects the relevant neural or computational dynamics (Mead 1990). Recent work has already explored neuromorphic and analog implementations of self-attention with the goal of reducing the computational and energetic cost of Transformer processing (Leroux, Finkbeiner, and Neftci 2025). From the present perspective, a further question is whether output-weight interconnections themselves can be physically realized as local, dynamically modifiable relations between processing components, rather than simulated indirectly through conventional matrix operations.

Such architectures could also bear on the scale required for present AI systems. Large language models achieve their capabilities with enormous amounts of static structure acquired through training, while dynamic transformations allow part of the processing structure to be constructed from the current input. This raises a more speculative question: could architectures that realize dynamic transformations more directly reduce the amount of static structure required to achieve comparable capabilities? The present analysis does not answer this question, but it identifies a functional principle whose alternative realizations may be relevant both to energy efficiency and to the development of smaller systems.

## 6.3 Biological Neural Processing

A third research direction concerns biological neural systems. The question is not whether the brain is a Transformer, but whether neural tissue can realize forms of dynamic processing functionally analogous to those identified above. At the mechanical-structural level, an output-weight interconnection requires a mechanism through which the activity of one component modifies how another connection transforms a signal. If biological neural systems contain relations of this kind, then at least some of the functional organization underlying SIDPP may be neurally realizable.

Axo-axonic synapses provide a relevant example. In such synapses, one axon contacts the presynaptic terminal of another and can modulate neurotransmitter release, thereby modifying the effect that the second connection has on its target (Cover and Mathur 2020). Local forms of interaction between neighboring synapses on the same dendrite can likewise modify synaptic efficacy in a context-dependent way (Weber et al. 2016). These mechanisms need not be regarded merely as functional analogues of output-weight interconnections. They leave open the possibility that the modulatory effect itself can be formally described as an operation in which the output of one synaptic interaction determines a variable parameter governing how another connection transforms its incoming signal. If such a description is adequate, the biological mechanism would realize an output-weight interconnection in a generalized sense, even though the operation by which the variable parameter acts on the incoming signal need not be ordinary multiplication. Figure 5 juxtaposes a neuromorphic realization of dynamic weights with an axo-axonic synapse as a possible biological analogue of output-weight modulation.

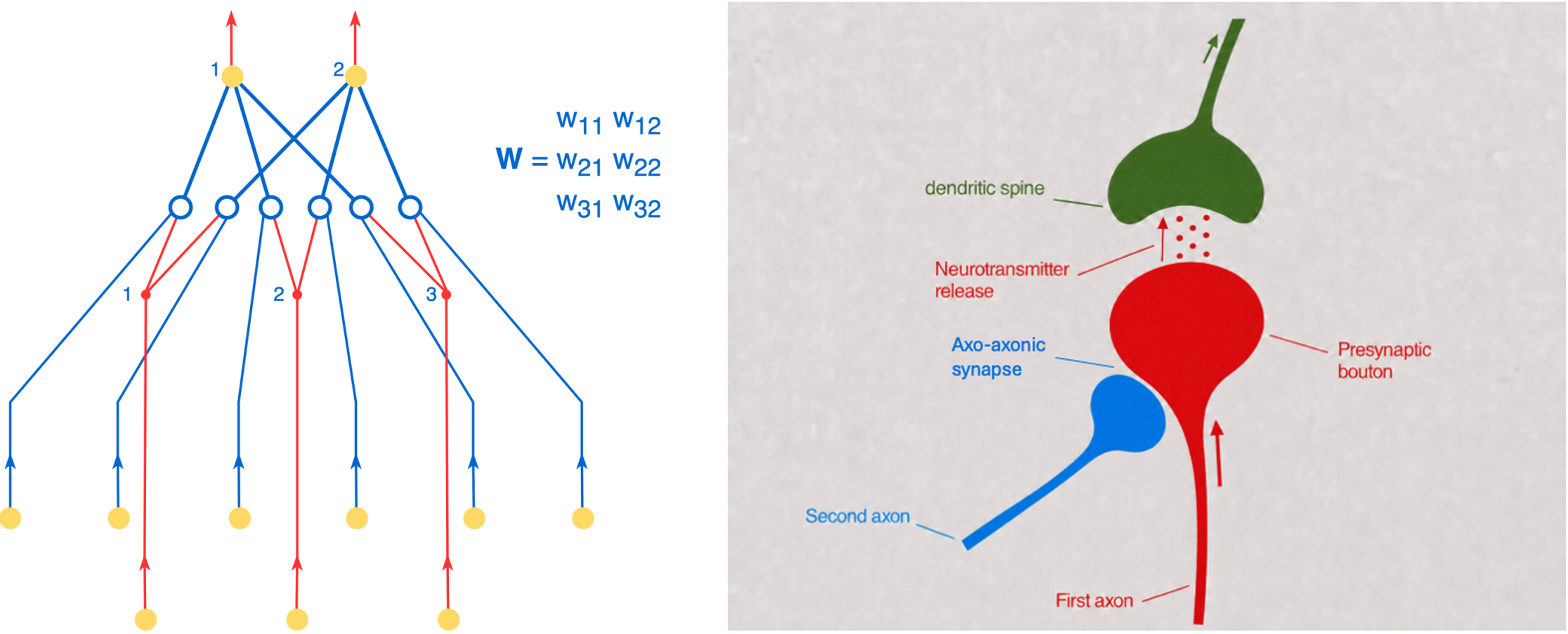

(a) (b)

**Figure 5.** A neuromorphic realization of dynamic weights and a possible biological analogue. (a) Neuromorphic realization of dynamic weights. (b) Axo-axonic synapse as a possible biological analogue of output-weight interconnections. Adapted from TR1, supplementary Figs. S19–S20.

The significance of these correspondences is therefore programmatic rather than identificatory. They show that the kind of local modulation required for dynamic transformations is compatible, in principle, with biological neural organization. This opens the possibility of investigating whether language processing in the brain depends on functional architectures that dynamically modify transformations on the basis of current activity, and whether similar principles could inform new neuromorphic AI systems. The resulting hypothesis is not that biological cognition reproduces the Transformer architecture, but that some of the processing principles made explicit by the Transformer may have more general neural realizations.

## 7 Conclusion

The Transformer revolution cannot be adequately understood simply as the result of more data, more parameters, and greater depth. These factors are important, and the Transformer remains genealogically and technically continuous with deep learning. The discontinuity identified in this paper concerns instead the organization of inference. The system does not merely apply transformations fixed through training: some of the transformations used during processing are constructed on the basis of the current input.

At the conceptual level, this amounts to adaptation before application. Static transforming concepts embody transformations acquired through training, while appropriate static transforming concepts are also used during inference to construct parameter structures that define dynamic transforming concepts. Previously acquired knowledge is therefore not only applied to the current input; part of it is first adapted to that input and only then applied. At the mechanical-structural level, the corresponding discontinuity is provided by output-weight interconnections, through which outputs of some networks become weights of others and thereby contribute to constructing the transformations those networks realize.

This organization gives rise to the mode of processing we have called Sequence-level Interactive Dynamic Parallel Processing (SIDPP). SIDPP combines parallel processing of token representations at the sequence level, interaction across those representations, and dynamically constructed transformations. The Transformer is one functional architecture that realizes this mode of processing, but SIDPP is not tied in principle to that architecture or to its standard implementation. This distinction is important because it turns the Transformer from a particular technological achievement into an exemplar of a more general processing principle.

Seen in this light, the Transformer shift is not merely a quantitative expansion of the statistical and deep-learning paradigm, but the emergence of a new paradigm based on a different principle of inference. This reinterpretation also opens further questions: how dynamic transformations can be analyzed to improve interpretability, predictability, and control; whether they can be realized more directly in alternative or neuromorphic architectures; and whether related forms of dynamic processing occur in biological neural systems. The conceptual and structural principles underlying the Transformer revolution may therefore be only beginning to be fully understood.